\documentclass[10pt,letterpaper]{article}
\usepackage{fullpage}
\usepackage{multirow}
\usepackage{graphicx}
\usepackage{amsmath}
\usepackage{amssymb}
\usepackage{appendix}
\usepackage[pagebackref=true,breaklinks=true,letterpaper=true,colorlinks,bookmarks=false]{hyperref}
\usepackage[linesnumbered,boxed]{algorithm2e}

\DeclareMathOperator*{\argmin}{argmin}
\DeclareMathOperator*{\argmax}{argmax}

\begin{document}
\title{Dual coordinate solvers for large-scale structural SVMs}
\author{Deva Ramanan\\ UC Irvine}
\date{}
\maketitle
This manuscript describes a method for training linear SVMs (including binary SVMs, SVM regression, and structural SVMs) from large, out-of-core training datasets. Current strategies for large-scale learning fall into one of two camps; batch algorithms which solve the learning problem given a finite datasets, and online algorithms which can process out-of-core datasets. The former typically requires datasets small enough to fit in memory. The latter is often phrased as a stochastic optimization problem \cite{bottou2008tradeoffs,shalev2007pegasos}; such algorithms enjoy strong theoretical properties but often require manual tuned annealing schedules, and may converge slowly for problems with large output spaces (e.g., structural SVMs). We discuss an algorithm for an ``intermediate'' regime in which the data is too large to fit in memory, but the active constraints (support vectors) are small enough to remain in memory. In this case, one can design rather efficient learning algorithms that are as stable as batch algorithms, but capable of processing out-of-core datasets. We have developed such a MATLAB-based solver and used it to train a series of recognition systems \cite{YangRF_CVPR2011,desai2012detecting,zhu2012face,hejrati2012analyzing,pirsiavash2012steerable,ren2013histograms,yang2013articulated,hejrati2014analysis,ghiasi2014parsing} for articulated pose estimation, facial analysis, 3D object recognition, and action classification, most of which has publicly-available code. This writeup describes the solver in detail.

{\bf Approach:} Our approach is closely based on data-subsampling algorithms for collecting hard examples~\cite{lsvm,lsvmj,dalal}, combined with the dual coordinate quadratic programming (QP) solver
described in liblinear \cite{fan2008liblinear}. The latter appears to be
current fastest method for learning linear SVMs. With regard to liblinear, we make two extensions (1) We generalize the solver to other types of
SVM problems such as (latent) structural SVMs (2) We modify it to behave as a partially-online algorithm, which only requires access to small amounts of in-memory data at a time. Data-subsampling algorithms typically operate by iterating between searching for hard examples and optimization over a batch of hard-examples in memory. With regard to these approaches, our approach differs in that (1) we use previously-computed solutions to``hot-start'' optimizations, making frequent calls to a batch solver considerable cheaper and (2) we track upper and lower bounds to derive an ``optimal'' schedule for exploring new data versus optimizing over the existing batch.

{\bf Overview:} Sec.~\ref{sec:gen} describes a general formulation of an SVM problem that encompasses many standard tasks such as multi-class classification and (latent) structural prediction. Sec.~\ref{sec:dual} derives its dual QP, and Sec.~\ref{sec:batch} describes a dual coordinate descent optimization algorithm. Sec.~\ref{sec:online} describes modifications for optimizing in an online fashion, allowing one to learn near-optimal models with a single pass over large, out-of-core datasets. Sec.~\ref{sec:theory} briefly touches on some theoretical issues that are necessary to ensure convergence. Finally, Sec.~\ref{sec:neg} and Sec.~\ref{sec:reg} describe modifications to our basic formulation to accommodate non-negativity constraints and flexible regularization schemes during learning.

\section{Generalized SVMs}
\label{sec:dual}
We first describe a general formulation of a SVM which encompasses various common problems such as binary classification, regression, and structured prediction. Assume we are given training data where the $i^{th}$ example is described by a set of $N_i$ vectors $\{x_{ij}\}$ and a set of $N_i$ scalars $\{l_{ij}\}$, where $j$ varies from 1 to $N_i$. We wish to solve the following optimization problem:
\begin{align}
  \argmin_w L(w) = \frac{1}{2}||w||^2 + \sum_{i} \max_{j \in N_i} (0,l_{ij} - w^T x_{ij}) \label{eq:opt}
\end{align}
We can write the above optimization as the following quadratic program (QP):
\begin{align}
  \label{eq:qp}
  &\argmin_{w, \xi \geq 0} \frac{1}{2}||w||^2 + \sum_i^N \xi_i\\
  &\text{s.t.} \qquad \forall i,j \in N_i \qquad  w^T x_{ij} \geq l_{ij} - \xi_i \nonumber
\end{align}
Eq. \eqref{eq:opt} and its QP variant \eqref{eq:qp} is a general form that encompasses binary SVMs, multiclass SVMs\cite{crammer2002algorithmic}, SVM regression \cite{smola2004tutorial}, structural SVMs \cite{taskar2003max,tsochantaridis2004svm} the convex variant of latent SVMs \cite{lsvm,lsvmj} and the convex variant of latent structural SVMs \cite{yu2009learning}. In Appendix~\ref{sec:gen}, we explicitly derive the various special cases. We will describe algorithms for handling large numbers of examples $i$, as well as (possibly exponentially) large values of $N_i$.

The above generalizes a traditional SVM learning formulation in several ways. Firstly, each example $x_{ij}$ comes equipped with its own margin $l_{ij}$. We shall see that this involves a relatively small modification to the QP solver. A more significant modification is that slack variables are now {\em  shared} across linear constraints. If each example $x_{ij}$ had its own slack variable $\xi_{ij}$, then \eqref{eq:qp} would be structurally equivalent to a standard SVM and amenable to standard QP optimization techniques. Intuitively, with independent slack variables, the set of examples $x_{ij}$ could contribute $N_i$ ``dollars'' to the loss. This could be a problem when $N_i$ is exponentially-large (as is the case for structured output spaces). By sharing slacks, the set of examples can only contribute at most one ``dollar'' to the loss. 

To further analyze the effect of shared slacks, we derive the dual QP by writing down the associated Lagrangian:
\begin{align}
L(w,\xi,\alpha,\mu) = \frac{1}{2}||w||^2 + \sum_i \xi_i - \sum_{ij} \alpha_{ij} (w \cdot x_{ij} - l_{ij} + \xi_i) - \sum_i \mu_i \xi_i
\end{align}

By strong duality
\begin{align}
\min_{w,\xi} \big[ \max_{\alpha \geq 0,\mu \geq 0} L(w,\alpha,\mu) \big]=\max_{\alpha \geq 0,\mu \geq 0} \big[ \min_{w,\xi}  L(w,\alpha,\mu) \big]
\end{align}

We take the derivative of the Lagrangian with respect to the primal
variables to get the KKT conditions:
\begin{align}
\frac{\partial{L(w,\alpha,\mu)}}{\partial{w}} =0 \quad \rightarrow \quad &w = \sum_{ij} \alpha_{ij} x_{ij} \label{eq:kkt}\\
\frac{\partial{L(w,\alpha,\mu)}}{\partial{\xi_i}} =0 \quad \rightarrow \quad &\sum_{j} \alpha_{ij} \leq 1 \quad \forall i
\end{align}

We write the dual of the QP in \eqref{eq:qp} as
\begin{align}
\label{eq:dual}
F(\alpha) = -&\frac{1}{2} \sum_{ij,kl} \alpha_{ij} x_{ij}^T x_{kl} \alpha_{kl} + \sum_{ij} l_{ij} \alpha_{ij}\\
\text{s.t.} \quad \forall i, \quad &\sum_j \alpha_{ij} \leq 1 \label{eq:lineq}\\
\forall i,j \in N_i, \quad &\alpha_{ij} \geq 0 \nonumber
\end{align}
We wish to maximize \eqref{eq:dual} over the dual variables $\alpha$. We can further analyze the nature of the optimal solution by considering {\em complementary slackness} conditions, which states that either a primal constraint is active, or its corresponding dual Lagrangian multiplier is 0:
\begin{align}
  \forall i,j \in N_i \quad {s.t} \quad \alpha_{ij} > 0, \quad w^T x_{ij} = l_{ij} - \xi_i
\end{align}

The above condition states that all examples with non-zero alpha (support vectors) associated with a single $i$ will incur the same slack loss $\xi_i$. In order words, these examples correspond to ``ties'' in the maximization over $\max_j(0, l_{ij} - w\cdot x_{ij})$ from \eqref{eq:opt}.  At the optimal dual solution, the linear constraint from \eqref{eq:lineq} delicately balances the influence of support vectors associated with $i$ to ensure they all pay the same slack loss. If each example $x_{ij}$ had its own slack variable $\xi_{ij}$, \eqref{eq:lineq} would be replaced by independent ``box constraints''  $\alpha_{ij} \leq 1$ for all $ij$. Box constraints are simpler to deal with because they decouple across the dual variables. Indeed, we show that the linear constraints considerable complicate the optimization problem.

\section{Batch optimization}
\label{sec:batch}
We now describe efficient training algorithms for solving dual QPs of the general form from \eqref{eq:dual}.  We begin by describing a solver that operates in a batch setting, requiring access to all training examples $i$ and all sets of constaints for each example $j \in N_i$. Our online algorithm will remove both these restrictions. The fastest current batch solver for linear SVMs appears to be liblinear \cite{fan2008liblinear}, which is a dual coordinate descent method. A naive implementation of a dual solver would require maintaining a $N \times N$ kernel matrix. The
innovation of liblinear is the realization that one can implicitly
represent the kernel matrix for linear SVMs by maintaining the primal weight
vector $w$, which is much smaller. We show a similar insight can be
used to design efficient dual solvers for generalized SVMs of the form from \eqref{eq:qp2}. Shared slacks considerably complicate coordinate-wise updates, but importantly, we describe an optimization schedule that is as fast as liblinear for much of the optimization.

To derive the modified optimization, let us first try to naively apply dual coordinate descent to optimizing \eqref{eq:dual}: let us pick a single dual variable $\alpha_{ij}$, and update it holding all
other $\alpha$'s fixed. This reduces to maximizing a 1-D
quadratic function subject to box constraints. This appears easy at first;
solve for the maximum of the quadratic and clip the solution to lie
within the box constraint. We solve for the offset $a$ that maximizes:


\begin{align}
\label{eq:coor}
F(\alpha + a\delta_{ij}) = &\frac{1}{2} h_{ij} a^2 + g_{ij}a + \text{constant}\\
\text{s.t.} \quad & 0 \leq  \alpha_i + a \leq 1 \nonumber
\end{align}

where $\alpha_i = \sum_j \alpha_{ij}$, $h_{ij} = -x_{ij}^Tx_{ij}$
(which can be precomputed), and the gradient
can be efficiently computed using $w$:
\begin{align}
g_{ij} &=  l_{ij} - \sum_{kl} x_{ij}^T x_{kl} \alpha_{kl} \nonumber
\\ &= l_{ij} - w^Tx_{ij}  \label{eq:grad} 
\end{align}

\begin{figure}[t!]
\centering
\includegraphics[width=\columnwidth]{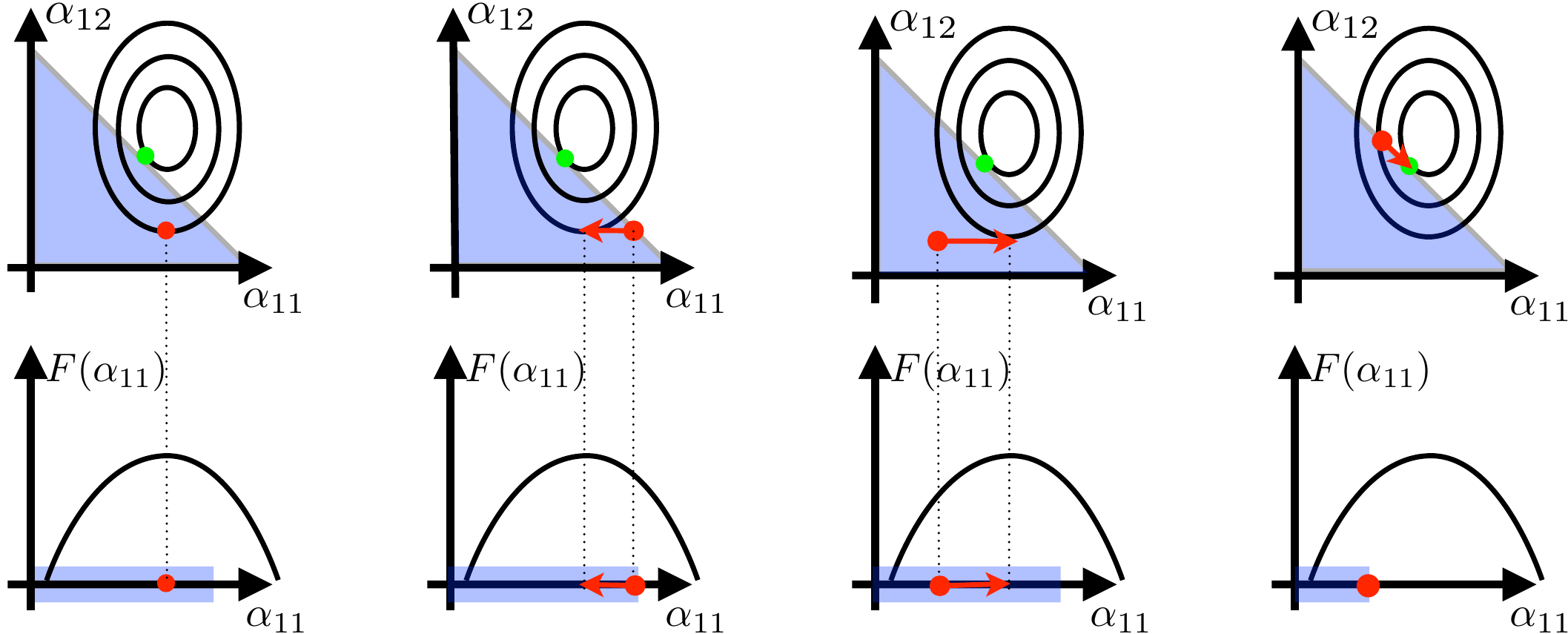}
(1) \hspace{110pt} (2) \hspace{110pt} (3) \hspace{110pt} (4)
\caption{We visualize various scenarios encountered during dual coordinate updates for a generalized SVM with two constraints that share slacks. On the top row, isocontours of the dual objective function $F(\alpha_{11},\alpha_{12})$ are drawn as ellipses, while the feasible region $\{\alpha_{11} \geq 0, \alpha_{11} \geq 0, \alpha_{11} + \alpha_{12} \leq 1\}$ is drawn in blue. For each scenario, the current value of $(\alpha_{11},\alpha_{12})$ is given by the red dot, while the optimal solution is given by the green dot. On the bottom, we show the one-dimensional quadratic problems corresponding to an update of $\alpha_{11}$ from the current solution. For scenarios {\bf (1),(2), and (3)}, $F$ can be improved by a coordinate-wise update of $\alpha_{11}$, which is easily found by computing the maximum of the one-dimensional quadratic and clipping the result to the blue feasible region \eqref{eq:solution1}. These updates are drawn as red arrows. For scenario {\bf(4)}, no coordinate-wise update can be made from the current solution because $F(\alpha_{12})$ is already at its optimum value given $\alpha_{11}$, and $F(\alpha_{12})$ cannot be further optimized because of the linear constraint ({\bf bottom}). Rather, both $\alpha_{11}$ and $\alpha_{12}$ need to be jointly updated ({\bf top}) as in \eqref{eq:solution2}.}
\label{fig:switch}
\end{figure}

There are four scenarios one can encounter when attempting to
maximize \eqref{eq:coor}, which are visually depicted in Fig.~\ref{fig:switch}:
\begin{enumerate}
\item $g_{ij}$ = 0, in which case $\alpha_{ij}$ is optimal for the
  current $\alpha$.
\item $g_{ij} < 0$, in which case decreasing $\alpha_{ij}$ will increase the dual.
\item $g_{ij} > 0$  and $\alpha_i < 1$, in which
  case increasing $\alpha_{ij}$ will increase the dual.
\item $g_{ij} > 0$ and $\alpha_i = 1$, in which case increasing $\alpha_{ij}$
  may increase the dual.
\end{enumerate}

For the first three scenarios, we can apply ``standard'' coordinate-wise updates that maximize \eqref{eq:coor} in
closed form:
\begin{align}
\label{eq:solution1}
  a^* = \min \Big( \max \big( -\alpha_i,\frac{g_{ij}} {h_{ij}}
  \big),1 - \alpha_i \Big)
\end{align}
\noindent yielding the update rule 
\begin{align} 
  a_{ij} := a_{ij} + a^* \label{eq:updateAij}
\end{align}
 For the last scenario, we would like to increase $\alpha_{ij}$ but
cannot because of the active linear constraint $\sum_j \alpha_{ij} =
1$. Let us select another dual variable ($\alpha_{ik}, k\neq j$) that shares this constraint to
possibly decrease. We would like to find the offset $a$ that maximizes:
\begin{align}
\label{eq:pair}
&F(\alpha + a\delta_{ij} - a\delta_{ik}) = \frac{1}{2}h'a^2 + g'a + \text{constant}\\
\text{s.t.} \quad &   0 \leq \alpha_{ij} + a \leq 1 \quad \text{and}
\quad 0 \leq \alpha_{ik} - a \leq 1 \nonumber
\end{align}
\noindent where $h' = h_{ij} + h_{ik} - 2x_{ij}^Tx_{ij}$ and $g' =
g_{ij} - g_{ik}$. Any value of $a$ will ensure that the
linear constraint is satisfied.
The above maximization can be computed in closed form:
\begin{align}
\label{eq:solution2}
 & a^* = \min \Big( \max \big( a_0,\frac{g_{ij}} {h_{ij}}
  \big),a_1 \Big) \quad \text{where} \\
a_0 = -&\max(\alpha_{ij},1-\alpha_{ik}) \quad \text{and} \quad  a_1=
\min(\alpha_{ik},1 - \alpha_{ij})  \nonumber
\end{align}
\noindent which yields the following coordinate updates:
\begin{align}
\alpha_{ij} := \alpha_{ij} + a^* \quad \text{and} \quad \alpha_{ik} :=
\alpha_{ik} - a^* \label{eq:updateAijk}
\end{align}

\subsection{Tracking $w$ and $\alpha_i$}
In order to enable efficient computation of the gradient for the next coordinate step, we track the change in $w$ using the KKT condition \eqref{eq:kkt}:
\begin{align}
w &:= w + a^*x_{ij} \quad \text{for single dual update; e.g., conditions 1-3 hold} \label{eq:updateW}\\
w &:= w + a^*(x_{ij} - x_{ik}) \quad \text{for pairwise dual update; e.g., condition 4 holds} \label{eq:updateWpair}
\end{align}
Similarly, we
can track the change in $\alpha_i$ from \eqref{eq:coor} :
\begin{align}
\alpha_i := \alpha_i + a^* \label{eq:updateAi}
\end{align}
\noindent which only needs to updated for the single dual update. 

\subsection{Dual coordinate-descent}
We provide pseudocode for our overall batch optimization algorithm below.

\begin{algorithm}[H]
\SetKwBlock{Loop}{Repeat}{}
\KwIn{$\{x_{ij},l_{ij}\}$}
\KwOut{w}
$\forall ij, \quad \alpha_{ij} := 0, \alpha_i := 0, w := 0$ \tcp*{Initialize variables (if not passed as arguments)}
\Loop {
  Randomly pick a dual variable $\alpha_{ij}$; \\
  Compute gradient $g_{ij}$ from \eqref{eq:grad};\\
  \uIf(\tcp*[f]{Find another variable if linear constraint is active}){$g_{ij} > \epsilon$ and $\alpha_i = 1$}
  {
    Randomly pick another dual variable with same $i$ ($\alpha_{ik}$ for $k \neq j$);\\
    Compute $a^*$ with \eqref{eq:solution2} ;\\
    Update $\alpha_{ij},\alpha_{ik},w$ with \eqref{eq:updateAijk},\eqref{eq:updateW}\\
  }
  \ElseIf(\tcc*[f]{Else update single dual variable})
{$|g_{ij}| > \epsilon$}
  {
    Compute $a^*$ with \eqref{eq:solution1};\\
   Update $\alpha_{ij},\alpha_i,w$ with \eqref{eq:updateAij}, \eqref{eq:updateAi},\eqref{eq:updateW}  
  }
}
\caption{$Optimize(\{x_{ij},l_{ij}\})$ performs batch optimization of a fixed dataset using multiple passes of dual coordinate descent. We also define a variant that can be ``hot-started'' with an existing set of dual variables, and optimized until some tolerance threshold $tol$ is met $Optimize(\{x_{ij},l_{ij},a_{ij}\},tol)$. \label{alg:batch}}
\end{algorithm}


{\bf Random sampling:} One may question the need for random sampling; why not iterate over dual variables ``in order''? The answer is that in practice, neighboring examples $x_{ij}$ will tend to be correlated (e.g., consider examples extracted from overlapping sliding windows in an image). In the extreme case, consider two identical training examples $x_1$ and $x_2$. After performing a dual update on $x_1$, $x_1$ will usually score better, and often pass the margin test under the newly-updated $w$. If we immediately visit $x_2$, it will also pass the margin test and $w$ will not be updated. However, assume we first visit an uncorrelated example (that does trigger $w$ to be updated) and then visit $x_2$. This allows us to effectively ``revisit'' $x_1$ in a single pass over our data. Hence a single (but randomly permuted) pass of coordinate descent effectively mimics multiple passes of (sequential) coordinate descent over correlated datasets. Lecun et al make similar observations to motivate random perturbations of data for stochastic gradient descent \cite{lecun2012efficient}.

{\bf Speed:}  In practice, we apply our randomized batch algorithm by performing a large number of sequential passes over random permutations of a fixed dataset.  During initial iterations, dual variables tend to be small and the linear inequality constraints in \eqref{eq:lineq} are not active. {\em This means that for much of our optimization, our solver updates a single dual variable at a time using \eqref{eq:updateAij}, making it essentially is fast as liblinear}. In later passes, the linear constraint tends to be active, and the solver is slower because update from \eqref{eq:updateAij} requires computing a dot product between two feature vectors with shared slack variables. In theory, one could cache these dot products in a reduced kernel matrix (since one needs to only store dot products between examples with shared slacks, rather than all $N^2$ examples). We found that computing them on-the-fly is still rather fast, and simplifies code.

{\bf Convergence:} With enough passes, the batch algorithm is guaranteed to converge (Sec.~\ref{sec:theory}). In the following, we provide a practical stopping criteria for convergence (within some tolerance). In the next section on online-learning, we will make use of such a tolerance to manage computation to repeated calls of a dynamic QP-solver. To define our stopping criteria, we closely follow the duality-based stopping criteria described in \cite{ocas}. Let $OPT = \min_w L(w)$, the optimal primal objective function value from \eqref{eq:opt}. Let us consider a candidate solution to the dual problem specified by a particular setting of dual variables $\alpha= \{\alpha_{ij}\}$. We can compute a lower bound on $OPT$ by with $F(\alpha)$, since all dual solutions are a lower bound on $OPT$ (by strong duality). We can also compute the associated primal weight vector $w(\alpha)=\sum_{ij} \alpha_{ij} x_{ij}$. We know that $L(w(\alpha))$ must be an upper bound on $OPT$, since $OPT$ is the minimal possible primal objective over all $w$:

\begin{align}
  &LB \leq OPT \leq UB \qquad \text{where}\\
LB &= F(\alpha) = -\frac{1}{2} w(\alpha)^Tw(\alpha) + l(\alpha) \label{eq:lb}\\
UB &= L(w(\alpha)) = \frac{1}{2} w(\alpha)^Tw(\alpha) + \sum_{ij} \max_j(0,l_{ij} - w(\alpha)^Tx_{ij}) \label{eq:ub}\\
w(\alpha) &= \sum_{ij} \alpha_{ij} x_{ij} \nonumber\\
l(\alpha) &= \sum_{ij} l_{ij} \alpha_{ij} \nonumber
\end{align}
It is straightforward to track changes to the lower bound from \eqref{eq:lb} by modifying lines 8 and 11 from Alg.~\ref{alg:batch} to maintain a running estimate of $l(\alpha)$ as dual variables are sequentially updated. The upper bound cannot be easily tracked, and instead has to be computed by passing over the entire set of data to compute the loss from \eqref{eq:ub} for a particular dual solution $\alpha$. In practice, one can simply update the upper bound occasionally, after a large number of dual updates. Once the upper and lower bound are found to lie within some tolerance $tol$, the batch algorithm terminates. Sec.~\ref{sec:theory} suggests that given sufficient iterations, the bounds must meet. The full interface to our batch algorithm is $Optimize(\{x_{ij},l_{ij},\alpha_{ij}\},tol)$ .

{\bf Approximate upper-bound:} We now describe an approximate upper bound that is easily tracked given a single sequential pass over a fixed dataset. Recall that our batch algorithm performs a large number of sequential passes over random permutations of our fixed dataset. The intuition behind our approximate upper bound is that we can use gradients $g_{ij}$, computed during dual updates from \eqref{eq:grad} to approximate the loss $l_{ij} - w(\alpha)^Tx_{ij}$: 
\begin{align}
UB' = \frac{1}{2} w(\alpha)^Tw(\alpha) + \sum_{ij} \max_j(0,g_{ij}(\alpha^t))
\end{align}
With some abuse of notation, we write $g_{ij}(\alpha^t)$ to explicitly denote the fact that gradients are computed with a changing set of dual variables at step $t$ of coordinate descent. If no dual variables are updated during a single sequential pass over the fixed dataset, then $\alpha^t = \alpha$ and the approximation is exact $UB' = UB$. In general we find that $UB' > UB$ since the loss due to a data example $l_{ij} - w(\alpha)^Tx_{ij}$ typically decreases after optimizing the dual variable associated with that data example. 
In practice, we keep track of this approximate upper bound, and whenever the tolerance criteria is satisfied with respect to this approximation, we compute the true upper-bound and perform another sequential pass if the true tolerance is not met. We find this speeds up batch optimization to convergence (up to $tol$) by factor of 2, compared to explicitly recomputing the true upper-bound after each sequential pass.

\section{Online learning} 
\label{sec:online}
In this section, we describe an efficient algorithm for online optimization, given large streaming datasets. Importantly, our online algorithm requires caching a small number of constraints $ij$, and so is appropriate for problems with large numbers of examples $i$ and/or problems with large numbers of shared constraints per example $j \in N_i$. The later is important for solving structured prediction problems, where $N_i$ may be exponentially large and pratically difficult to explicitly enumerate. When applied in a cyclic fashion on finite datasets, our online algorithm is garuanteed to converge to the optimal solution.

{\bf Loss-augmented inference:}  To deal with exponentially-large $N_i$, our online algorithm does {\em not} require access to the entire set of constraints $\{x_{ij}, j \in N_j\}$ associated with example $i$, but rather assumes that this set can be efficiently searched. Specifically, we assume the user provides a function that, given a current weight vector $w$ and example $i$, computes the ``worst-offending'' constraint $j \in N_i$. This is often known as a ``loss-augmented inference'' problem:
\begin{align}
 \text{WorstOffender}_i &= \argmax_{j \in N_i} \big[ l_{ij} - w^Tx_{ij} \big]\\
&= \argmax_j g_{ij} \quad \text{where $g_{ij}$ is defined as \eqref{eq:grad}} \label{eq:loss}
\end{align}
The above assumption is often quite reasonable as a similar maximization problem $(\max_j w^T x_{ij})$ must typically be solved for inference at test-time ({\em e.g.}, for structured prediction). 
We make two additional notes. Firstly, the $\max$ corresponding to the above $\argmax$ is the loss due to example $i$ for the current model $w$, which is required to to compute $L(w)$ from \eqref{eq:opt}. Secondly, the worst-offender is equivalent to the constraint $j$ with largest gradient $g_{ij}$. This property can be used to to efficiently select ``good'' constraints for coordinate updates, that are likely to increase the dual objective.

{\bf One-pass coordinate-descent:} The batch algorithm from Alg.\ref{alg:batch} can be ``trivially'' turned into a online algorithm by performing a single, sequential pass over a streaming dataset. Given a new data point $i$, one can sample a random $j \in N_i$ (or select $j$ with the largest gradient, through loss augmented inference as described above) and compute its associated dual variable $\alpha_{ij}$. 
Because this is first time example $i$ is encountered, its associated linear constraint $\sum_{ij} \alpha_{ij} \leq 1$ is not active. This means one can apply a simple coordinate-wise update of $\alpha_{ij}$, {\em without} the need to consider pairs of variables. This online algorithm has two notable properties. (1) This optimization step is guaranteed to not decrease the dual objective function. This is in contrast to online algorithms such as the perceptron or stochastic gradient descent that may take steps in the wrong direction. (2) The algorithm never requires the
computation of the kernel matrix, and instead maintains an estimate of
the primal variable $w$. This means that the storage requirement is constant with respect to the number of training examples, rather than quadratic (as required for a kernel matrix). 

{\bf Exploration vs optimization:} 
A crucial question in terms of convergence time is the order of
 optimization of the dual variables $\alpha_{ij}$. The above one-pass sequential algorithm takes an online perspective, where one continually explores new training examples. As $w$ is being learned, at some point many (if not most) new examples will be ``easy'' and pass the margin test. For such cases, $\alpha_{ij} = 0$ and $g_{ij} \leq 0$, implying that the given dual coordinate step does not trigger an update, making learning inefficient. The LaRank algorithm \cite{bordes2007solving} makes the observation that it is beneficial to frequently revisit {\em past} examples with a non-zero alpha, since they are more likely to trigger an update to their dual value. Previously examples that looked ``hard'' often tend to remain ``hard''. This can be implemented by maintaining a cache of ``hard examples'' \cite{lsvmj}, or support vectors and routinely optimizing over them while exploring new examples. This is basis for much of the literature on both batch and online SVMs, through the use of heuristics such as active-set selection~\cite{fan2008liblinear,tsochantaridis2004svm} and new-process/optimization steps~\cite{bordes2007solving}. We find that the precise scheduling of the optimization over new points (which we call {\em exploration}) versus existing support vectors (which we call {\em optimization}) is crucial. 

{\bf Scheduling strategies:} One scheduling strategy is to continually {\em explore} a new data point, analogous to the one-pass algorithm described above. Instead, LaRank suggests a exploring/optimization ratio of 10:1; revisit 10 examples from the cache for every new training example. Still another popular approach is to {\em explore} new data points by adding them to cache up to some fixed memory limit, and then {\em optimize} the cache to convergence and repeat. The hard-negative mining solver of \cite{lsvmj} does exactly this. We make a number of observations to derive our strategy. First, successive calls to a dual solver can be hot-started from the previous dual solution, making them quite cheap. Secondly, it is advantageous to behave like an online algorithm ({\em explore}) during initial stages of learning, and behave more batch-like ({\em optimize}) during later stages of learning when the model is close to convergence. We propose here a novel on-line cutting plane algorithm \cite{joachims2006tls,Joachims/etal/09a} that maintains running estimates of lower and upper bounds of the primal objective that naturally trade off the exploration of new data points versus the optimization of existing support vectors.

{\bf Online duality-gap:} At any point in time during learning, we have a cache of examples and associated dual variables in
memory $\{x_{ij},l_{ij}, \alpha_{ij}: ij \in A\}$. Together, these completely specify a primal weight vector $w(\alpha) = \sum_{ij \in A} \alpha_{ij}x_{ij}$. We must decide between two choices; we can either further optimize $\alpha$ over the current examples in memory, or we can query for a brand new, unseen datapoint. From one perspective, the current cache of examples specifies a well-defined finite QP, and we may as well optimize that QP to completion. This would allow us to define a good $w$ that would, in-turn, allow us to collect more relevant hard examples in the future. From another perspective, we might always choose to optimize with respect to new data ({\em explore}) rather than optimize over an example that we have already seen. We posit that one should ideally make the choice that produces the greatest increase in dual objective function value. Since it is difficult to compute the potential increase, we adopt the following strategy: we always choose to {\em explore} a new, unseen data point, unless the duality gap for the current QP solution is large (above $tol$), in which case we would rather {\em optimize} over the current cache to reduce the duality gap. We efficiently track the duality gap in an online fashion with the following streaming algorithm:

\begin{algorithm}[H]
\KwIn{Streaming dataset $\{x_{ij},l_{ij}\}$ and $tol$}
\KwOut{w}
$A = \{\}$;\\ $w :=0$; $UB := 0$; $LB := 0$ \tcp*{Initialize variables}
\For{i = $1:\infty$}
{
  Consider new example $x_i =\{x_{ij}: j \in N_i\}$ \tcp*{Explore new example}
  Compute $j$ with maximum gradient $\max_j g_{ij}$ from \eqref{eq:grad}; \tcp*{Find ``worst offender''}
  \If(\tcp*[f]{Add to cache if it violates margin}){$g_{ij} > 0$} 
  {
    $UB := UB + g_{ij}$; \\
    $\alpha_{ij} := 0$;\\
    $A := A \cap (ij)$ 
  }

  \If(\tcp*[f]{Optimize over cache if duality gap is violated}){UB - LB $>$ tol}
  {
    $(w,\alpha_A,LB,UB)$ = Optimize$(\{x_A,l_A,\alpha_A\},tol)$;\\
    $A := A \setminus \{ij: \alpha_{ij} = 0\}$ \tcp*{(Optionally) remove non-support-vectors from cache}
  }
}
\caption{The above online algorithm performs one pass of learning across a dataset by maintaining a cache of examples indices $A=\{(ij)\}$. At any point in time, the associated dual variables $\{\alpha_{ij}\}$ encode the optimal model $w$ for the QP defined by the cached examples, up to the duality gap $tol$. 
\label{alg:online}}
\end{algorithm}


{\bf Convergence:} To examine the convergence of the online algorithm, let us make a distinction between two QP problems. The cached-QP is the QP defined by the current set of examples in cache $A$. The full-QP is the QP defined by the possibly infinitely-large set of examples in the full dataset. During the online algorithm, UB and LB are always upper and lower bounds on the cached-QP. LB is also a lower bound on the full-QP. One can derive this fact by setting dual variables for all examples not in the cache to $0$, and scoring the resulting full-$\alpha$ vector under the dual objective, which must be $F(\alpha) = LB$. Crucially though, $UB$ is {\em not} an upper bound on the full-QP. This makes intuitive sense; it is hard to upper bound our loss without seeing all the data. Hence convergence for the online algorithm cannot be strictly guaranteed. However, if we apply the online algorithm by cycling over the dataset multiple times, one can ensure convergence with similar arguments to our batch optimization algorithm. Moreover, after learning a weight vector $w(\alpha_A)$, we can verify that it is optimal by computing a true upper bound $L(w(\alpha_A))$. This can be computed with a single, out-of-core pass over the entire dataset. We find that in practice, a single pass through large datasets often suffices for convergence (where if desired, convergence can be explicitly verified with an additional single pass).

{\bf Pruning:} The last step of our online algorithm includes an optional pruning phase that removes non support vectors from the cache. It is crucial to note that pruning does not affect the lower-bound $LB$, since the dual objective $F(\alpha)$ will not change by removing constraints with zero $\alpha$ values. Importantly, this lower-bound still holds even if the current cache is {\em not} optimized to convergence. Hence overall convergence (with cyclic online passes) is still guaranteed. However, it is possible that a pruned constraint later becomes a support vector after future updates to $w$, making the optimization somewhat inefficient and slowing convergence. Other active set methods make use of various heuristics for pruning a constraint, such as waiting until it remains inactive for a fixed number (50) of iterations \cite{Joachims/00c}. For large datasets, we found it useful to aggressively prune; one a constraint appears easy, immediately remove it from the cache. Our intuition is that large datasets contain many correlated examples, so the solver will encounter a similar example later on during online optimization.

\section{Theoretical guarantees}
\label{sec:theory}
In this section, we briefly point to some theoretical analysis that is necessary to ensure to show that the batch and cyclical-online version of our algorithm will converge to the global optimum. In particular, this analysis reveals that convergence of coordinate-wise updates does {\em not} necessarily produce the globally optimal solution, motivating the need for optimizing pairs of dual variables in Alg.~\ref{alg:batch}.

{\bf Global optimality:} It is straightforward to define the optimality conditions of differentiable, unconstrained convex functions - the gradient of the function must be zero. Our dual objective function is differentiable, but constrained \eqref{eq:dual}. Boyd shows that optimality conditions in such cases is more subtle~\cite{boyd2004convex}. Specifically, $\alpha$ is optimal if and only if there is no {\em feasible descent direction}. Or in other words, it suffices to show that any small step along any direction that is feasible (in the convex set defined by our constraints) increases the value of the dual objective. \cite{bordes2005fast} prove that is suffices to show that no improvement can be made along a set of ``witness'' directions that form a basis for the feasible set of directions.


{\bf Joint  optimization:} If no constraints are active at a given $\alpha$, the coordinate axes do define a basis set. This means that for examples $i$ for which the linear constraint $\sum_{j} \alpha_{ij} \leq 1$ is not active, it suffices to ensure that the dual cannot be improved by independently perturbing each dual variable $\alpha_{ij}$. However, this is not true for examples $i$ with active linear constraints; its possible that no improvement can be made by taking a step along any dual variable, but an improvement {\em can} be made along a direction that changes a pair of dual variables (Fig.~\ref{fig:switch}-(4)). This necessitates the need for the switch clauses in Alg.~\ref{alg:batch} that enumerate possibly active constraints, and precisely the reason why shared slacks in a structural SVM require joint optimization over pairs of dual variables.

{\bf Cyclic optimization:} Consider an algorithm that randomly samples update directions with any distribution such that all feasible directions can be drawn with non-zero probability; \cite{bordes2005fast} show that such an algorithm probably convergences to the optimum, within some specified tolerance, in finite time. Our batch algorithm and cyclic variant of our online algorithm satisfy this premise because they consider directions along each dual variable, as well as linear combinations of linearly-constrained variables.

\section{Non-negativity constraints}
\label{sec:neg}
We now describe a simple modification to our proposed algorithms that accept non-negativity constraints on parameters $w$:
\begin{align}
\label{eq:qpnn}
 L(w,\xi) = &\frac{1}{2}||w||^2 + \sum_i \xi_i\\
\text{s.t.} \quad &w^T x_{ij} > l_{ij} - \xi_i \nonumber\\
& \xi_i \geq 0 \nonumber\\
& w_k \geq0 \quad \forall k \in N \label{eq:nn}
\end{align}
The above is equivalent to our original formulation from \eqref{eq:qp} with an additional set of non-negativity constraints for a subset of parameters $w_k$ \eqref{eq:nn}. The associated Lagrangian is
\begin{align}
L(w,b,\xi,\alpha,\mu,\beta_k) = \frac{1}{2}||w||^2 + \sum_i \xi_i - \sum_{ij}
\alpha_{ij} (w \cdot x_{ij} - l_{ij} + \xi_i) -
\sum_i \mu_i \xi_i - \beta \cdot w
\end{align}
To simplify notation, we have assumed that all parameters are constrained to be non-negative in the above Lagrangian, but our final algorithm will allow for an abitrary subset $N$. By strong duality
\begin{align}
\min_{w,b,\xi} \big[ \max_{\alpha \geq 0,\mu \geq 0, \beta \geq 0} L(w,b,\alpha,\mu) \big] =\max_{\alpha \geq 0,\mu \geq 0, \beta \geq 0} \big[ \min_{w,b,\xi}  L(w,b,\alpha,\mu) \big]
\end{align}

We take the derivative of the Lagrangian with respect to the primal
variables to get the KKT conditions:
\begin{align}
&w = \sum_{ij} \alpha_{ij} x_{ij} + \beta\\
&\sum_{j} \alpha_{ij} \leq 1 \quad \forall i
\end{align}

We can write the dual of the QP in \eqref{eq:qpnn} as
\begin{align}
\label{eq:dualnn}
F(\alpha,\beta) = -&\frac{1}{2} ||\sum_{ij} \alpha_{ij} x_{ij} + \beta||^2 + \sum_{ij} l_{ij} \alpha_{ij}\\
\text{s.t.} \quad &\sum_j \alpha_{ij} \leq 1 \nonumber\\
&\alpha_{ij} \geq 0 \nonumber
\end{align}

We iterate between optimizing a single $\alpha_i$ holding $\beta$ fixed, followed by optimizing $\beta$. One can show this is equivalent to zero-ing our negative parameters during dual updates:

\begin{enumerate}
\item Update $\alpha_{ij},w$ with a coordinate descent update.
\item Update $\beta$ by $w[k] = \max(w[k],0), \forall k \in \{1 \ldots N \}$.
\end{enumerate}

\section{Flexible regularization}
\label{sec:reg}
This section will describe a method for using the aforementioned solver to solve a more general SVM problem with a Gaussian regularization or ``prior'' on $w$ given by $(\mu,\Sigma)$:
\begin{align}
\label{eq:qp-prior}
 \argmin_{w,\xi} &\frac{1}{2}||(w-w_0)R||^2 + \sum_i \xi_i\\
\text{s.t.} \quad &w^T x_{ij} > l_{ij} - \xi_i \nonumber\\
& \xi_i \geq 0 \nonumber
\end{align}

where $w_0 = \mu$, and $R = \Sigma^{-1/2}$. We can massage \eqref{eq:qp-prior} into \eqref{eq:qp} with the substitution ${\hat w} = (w-w_0)R$:

\begin{align}
\label{eq:primal}
 \argmin_{{\hat w},\xi} &\frac{1}{2}||{\hat w}||^2 + \sum_i \xi_i\\
\text{s.t.} \quad &{\hat w}^T {\hat x}_{ij} > {\hat l}_{ij} - \xi_i \nonumber\\
& \xi_i \geq 0 \nonumber\\
\text{where } & {\hat w} = (w-w_0)R \nonumber\\
& {\hat x}_{ij} = R^{-1} x_{ij} \nonumber \\
& {\hat l}_{ij} = l_{ij} - w_o \cdot x_{ij} \nonumber
\end{align}

We assume that $\Sigma$ is full rank, implying that $R^{-1}$ exists. An important special case is given by a diagonal matrix $\Sigma$, which corresponds to an arbitrary regularization of each parameter associated with a particular feature. This is useful, for example, when regularizing a feature vector constructed from heterogeneous features (such as appearance features, spatial features, and biases). After solving for the re-parametrized weight vector ${\hat w}$ by optimizing the QP from \eqref{eq:primal}, one can recover the score of the original weight vector with the following:
\begin{align}
w \cdot x_{ij} = ({\hat w} + w_0R) \cdot {\hat x}_{ij}  
\end{align}

\section{Conclusion}

We have described a dual coordinate solver for solving general SVM problems (including multiclass, structural, and latent variations) with out-of-core, or even streaming datasets. The ideas described here are implemented in publicly available solvers released in \cite{YangRF_CVPR2011,desai2012detecting,zhu2012face,hejrati2012analyzing,ren2013histograms}. 

{\bf Acknowledgements:} Funding for this research was provided by NSF Grant 0954083 and ONR-MURI
Grant N00014-10-1-0933. We also gladly acknowledge co-authors for numerous discussions and considerable debugging efforts.

\begin{appendices}
  \section{Generalized SVMs}
  \label{sec:gen}
  In this section, we show that various SVM problems can be written as general cases of our underlying QP \eqref{eq:qp}. This allows our optimization algorithms and solver to be applicable to a wide range of problems, including binary prediction, regression, structured prediction, and latent estimation. For conciseness, we write the QP from \eqref{eq:qp} here:
\begin{align}
  \label{eq:qp2}
  &\argmin_{w, \xi \geq 0} \frac{1}{2}||w||^2 + \sum_i^N \xi_i\\
  &\text{s.t.} \qquad \forall i,j \in N_i \qquad  w^T x_{ij} \geq l_{ij} - \xi_i \nonumber
\end{align}

  \subsection{Binary classification}
  A linearly-parametrized binary classifier predicts a binary label for an input $x$:
  \begin{align}
    \text{Label}(x) = \{ w^T x > 0 \}
  \end{align}
  The associated learning problem is defined by a dataset of labeled examples $\{x_i,y_i\}$, where $x_i \in \mathcal{R}^N, y_i \in \{-1,1\}$:
  \begin{align}
    \quad &\argmin_{\beta} \frac{1}{2}||\beta||^2 + C \sum_i \xi_i \label{eq:svm}\\
    &\text{s.t.} \quad y_i(\beta^Tx_i + b) \geq 1 - \xi_i \nonumber
  \end{align}
  One can convert the above into \eqref{eq:qp} with the following: first append a constant value $v$ to each feature to model the bias term with $x'_i = (x_i,v)$ where $v=1$ is the typical choice. This allows us to write $\beta^Tx_i + b = w^T x'_i$ where $w = (\beta,b)$. We then multiply in the class label $y_i$ and slack scaling $C$ into each feature $x'$, yielding $x_{ij} = (Cy_ix_i,Cy_i)$, where $j \in \{1\}$, $N_i = 1$ and  $l_{ij} = C$.
  
  {\bf Bias term:} The above mapping does not precisely correspond to \eqref{eq:qp} because the bias $b$ is now regularized. This means the learning objective function will prefer biases closer to 0, while \eqref{eq:svm} does not favor any particular bias. This may or may not be desired. In practice, one can decrease the effect of regularization by appending a large constant value $v$. For example, a model learned with $v=100$ will tend to produce larger effective biases $b$ than $v=1$. In the limit that $v \rightarrow \infty$, \eqref{eq:svm} does map directly to \eqref{eq:qp}. In Sec.~\ref{sec:reg}, we describe a modification to \eqref{eq:qp} that allows for arbitrary (but finite) regularization factors for individual parameters. For the description of subsequent SVM problems, we omit the bias term for notational simplicity.

{\bf Margin-rescaling:} The above formulation can easily handle example-specific margins: for example, we may require that certain ``prototypical'' positives score higher than 2 rather than the standard margin of 1. This can be done by defining $l_{ij} = \text{margin}_i$, where $\text{margin}_i$ is the margin associated with example $i$. In the literature, this modification is sometimes known as margin-rescaling. We will see that margin rescaling is one core component of structural SVM problems.

{\bf Cost-sensitive examples (slack-rescaling):} The above formulation can be easily extended to cost-sensitive SVMs by defining $l_{ij} = C_i$ and $x_{ij} = (C_iy_ix_i,C_iy_i)$. This is sometimes known as slack rescaling.  For example, one could define a different cost penalty for positive versus negative examples. Such class-specific costs have been shown be useful for learning classifiers from imbalanced datasets~\cite{akbani2004applying}.  For the description of subsequent SVM problems, we omit any slack rescaling term ($C$ or $C_i$) for notational simplicity, though they can always be incorporated by scaling $x_{ij}$ and $l_{ij}$.

\subsection{Multiclass SVMs} A linearly-parametrized multiclass predictor produces a class label for $x$ with the following:
$$\text{Label}(x) = \argmax_{j \in \{1 \ldots K\}} w_j^Tx$$
The associated learning problem is defined by a dataset $\{x_i,y_i\}$ where $x_i \in \mathcal{R}^N$ and  $y_i \in \{1,2,\ldots K\}$. There exist many approaches to multiclass prediction that reduce the problem to a series of binary prediction problems (say, by training $K$ 1-vs-all predictors or $K^2$ pairwise predictors). Each of the core binary prediction problems can be written as \eqref{eq:svm}, and so can be directly mapped to \eqref{eq:qp}. Here, we describe the particular multiclass formulation from \cite{crammer2002algorithmic} which requires an explicit mapping:
\begin{align}
\quad &\argmin_{w,\xi \geq 0} \frac{1}{2} \sum_j ||w_j||^2 + \sum_i \xi_i\\
& \text{s.t.} \forall i,j \neq y_i  \quad w_{y^i}^Tx_i - w_j^T x_i \geq \text{loss}(y_i,j) - \xi_i
\end{align}
The above formulation states that for example $i$, the score of the true class $y_i$ should dominate the score of any other class $j$ by $\text{loss}(y_i,j)$; if not, we should pay the difference (the slack). For example, given a multi-class problem where the class labels are car, bus, and person, one may wish to penalize mistakes that label a car as a person higher than those that label a car as a bus. In the general setting, this can be specified with a loss function $\text{loss}(y_i,j)$ that specifies the cost of labeling class $y_i$ as class $j$. The original formulation from \cite{crammer2002algorithmic} defined a 0-1 loss where $\text{loss}(y_i,j) = 0$ for $j = y_i$ and $\text{loss}(y_i,j) = 1$ for $j \neq y_i$. Finally, we have omitted an explicit class-specific bias term $b_j$ in the above formulation, but one can apply the same trick of appending a constant value to feature $x_i$.

The above form can be massaged into \eqref{eq:qp} by the following: let us define $w = (w_1,\ldots,w_K)$ as a $NK$-long vector of concatenated class-specific weights $w_j$, and $\phi(x_i,j)$ as a $NK$-length sparse vector with $N$ non-zero entries corresponding to the interval given by class $j$. These two definitions allow us to write $w_j^Tx_i = w^T\phi(x_i,j)$. This in turn allows us to define $x_{ij} = \phi(x_i,y_i) - \phi(x_i,j)$ , which then maps the above into \eqref{eq:qp}, where $N_i = (K-1)$.

\subsection{Structural SVMs}
A linearly-parametrized structural predictor produces a label of the form
$$\text{Label}(x) = \argmax_{y \in Y} w^T \phi(x_i,y)$$
where $Y$ represents a (possibly exponentially-large) structured output space. The associated learning problem is given by a dataset $\{x_i,y_i\}$ where $x_i \in \mathcal{R}^N$ and $y_i \in Y$:
\begin{align}
\quad &\argmin_{w,\xi \geq 0} \frac{1}{2} ||w||^2 + \sum_i \xi_i\\
& \text{s.t.} \forall i,{h \in Y}  \quad w^T\phi(x_i,y_i) - w^T \phi(x_i,h) \geq \text{loss}(y_i,h) - \xi_i
\end{align}

One can define $N_i = |Y|, x_{ij} = \phi(x_i,y_i) - \phi(x_i,j)$ and $l_{ij} = \text{loss}(y_i,j)$, where $j = h$ is interpreted as an index into the output space $Y$. 

\subsection{Latent SVMs} A latent SVM produces a binary prediction by searching over a latent variable
\begin{align}
\text{Labe}l(x) = \{ \max_{z \in Z} w \cdot(x,z) > 0 \}
\end{align}
where $Z$ represents a (possibly exponentially-large) latent space. In latent-SVM learning, and in particular, the convex optimization stage of coordinate descent \cite{lsvm}, each training example is given by $\{x_i,z_i,y_i\}$ where $y_i \in \{-1,1\}$, and $z_i$ are latent variables specified for positive examples:
\begin{align}
&\argmin_{w,\xi \geq 0} \frac{1}{2} ||w||^2 + \sum_i \xi_i\\
\text{s.t.} &\forall i \in \text{pos} \qquad w^T \phi(x_i,z_i) \geq 1 - \xi_i\\
\text{s.t.} &\forall i \in \text{neg}, g \in Z \qquad w^T \phi(x_i,g) \leq -1 + \xi_i
\end{align}
 One can map this to the above problem with the following: for $i \in \text{pos}$, $N_i = 1, x_{ij} = \phi(x_i,z_i), l_{ij} = 1$. For $i \in \text{neg}$, $N_i = |Z|$, $x_{ij} = -\phi(x_i,j), l_{ij} = -1$ where $j = g$.

\subsection{Latent structural SVMs}
One can extend the above model to the latent structural case, where the predictor behaves as follows:
$$ \text{Label}(x) = \argmax_{y \in Y} \big[ \max_{z \in Z} w^T \phi(x,y,z) \big]$$
The associated learning problem is defined by a dataset $\{x_i,z_i,y_i\}$ where $y_i \in Y$ is a structured label rather than a binary one. In this scenario, the analogous convex step of ``coordinate descent'' corresponds the optimizing the following optimization:
\begin{align}
&\argmin_{w,\xi \geq 0} \frac{1}{2} ||w||^2 + \sum_i \xi_i\\
\text{s.t.} \forall i, h \in Y, g \in Z,  &\qquad w^T\phi(x_i,y_i,z_i) - w^T \phi(x_i,h,g) \geq \text{loss}(y_i,h,g) - \xi_i \nonumber
\end{align}
This can be mapped to our general formulation by defining $N_i = |Y| |Z|$, $x_{ij} = \phi(x_i,y_i,z_i) - \phi(x_i,j)$ for $j \in Y \times Z$.

\subsection{Regression}
A linear regressor makes the following predictions
$$\text{Label}(x) = w^Tx $$
The associated SVM regression problem is specified by a dataset $\{x_i,y_i\}$ where $x_i \in \mathcal{R}^N$ and  $y_i \in \mathcal{R}$:
\begin{align}
\argmin_{w,\xi \geq 0} &\frac{1}{2} ||w||^2 + \sum_i (\xi_i + \xi_i^*)\\
\text{s.t.} \forall i,  \quad &w^Tx_i \geq y_i - \epsilon - \xi_i \nonumber\\
&w^Tx_i \leq y_i + \epsilon + \xi^*_i \nonumber
\end{align}
The above constraints can be converted to the from \eqref{eq:qp} by doubling the number of constraints by defining $(x'_i,y'_i) = (x_i,y_i - \epsilon)$ and $(x'_{2i},y'_{2i}) = (-x_i,-y_i - \epsilon)$ and $N_i = N_{2i} = 1$.

{\bf Summary:} In this section, we have shown that many previously-proposed SVM problems can be written as instances of the generic problem in \eqref{eq:opt}, which can be written as the quadratic program (QP) in \eqref{eq:qp}. 

\end{appendices}

{\small
  \bibliographystyle{plain}
  \bibliography{refs.bib}
}

\end{document}